%% file: 00_main.tex
\documentclass[10pt,twocolumn,letterpaper]{article}

\usepackage{cvpr}   

\usepackage{graphicx}
\usepackage{amsmath}
\usepackage{amssymb}
\usepackage{booktabs}
\usepackage{multirow}
\usepackage{booktabs}

\usepackage{algorithm}
\usepackage{algorithmicx}
\usepackage{mathtools}
\usepackage{adjustbox}
\usepackage{import}
\usepackage[noend]{algpseudocode}
\usepackage{caption}
\usepackage{subcaption}
\usepackage[dvipsnames]{xcolor}

\usepackage[pagebackref,breaklinks,colorlinks]{hyperref}

\usepackage[capitalize]{cleveref}
\crefname{section}{Sec.}{Secs.}
\Crefname{section}{Section}{Sections}
\Crefname{table}{Table}{Tables}
\crefname{table}{Tab.}{Tabs.}

\def\cvprPaperID{*****} 
\def\confName{CVPR}
\newcommand{\ours}{PHALP }
\begin{document}

\title{Tracking People by Predicting 3D Appearance, Location \& Pose}

\author{Jathushan Rajasegaran, Georgios Pavlakos, Angjoo Kanazawa, Jitendra Malik\\
UC Berkeley
}
\maketitle

\input{01_abstract}
\input{02_introduction}
\input{03_related_work}
\input{04_technical_approach}
\input{05_experiments}
\input{06_conclusion}

\newpage
{\small
\bibliographystyle{ieee_fullname}
\bibliography{egbib}
}

\end{document}

%% file: 01_abstract.tex
\begin{abstract}
In this paper, we present an approach for tracking people in monocular videos, by predicting their future 3D representations. To achieve this, we first lift people to 3D from a single frame in a robust way. This lifting includes information about the 3D pose of the person, his or her location in the 3D space, and the 3D appearance. As we track a person, we collect 3D observations over time in a tracklet representation. Given the 3D nature of our observations, we build temporal models for each one of the previous attributes. We use these models to predict the future state of the tracklet, including 3D location, 3D appearance, and 3D pose. For a future frame, we compute the similarity between the predicted state of a tracklet and the single frame observations in a probabilistic manner. Association is solved with simple Hungarian matching, and the matches are used to update the respective tracklets. We evaluate our approach on various benchmarks and report state-of-the-art results.
\end{abstract}

%% file: 02_introduction.tex
\section{Introduction}
\label{sec:intro}

\begin{figure}[!t]
    \centering
    \includegraphics[width=0.40\textwidth]{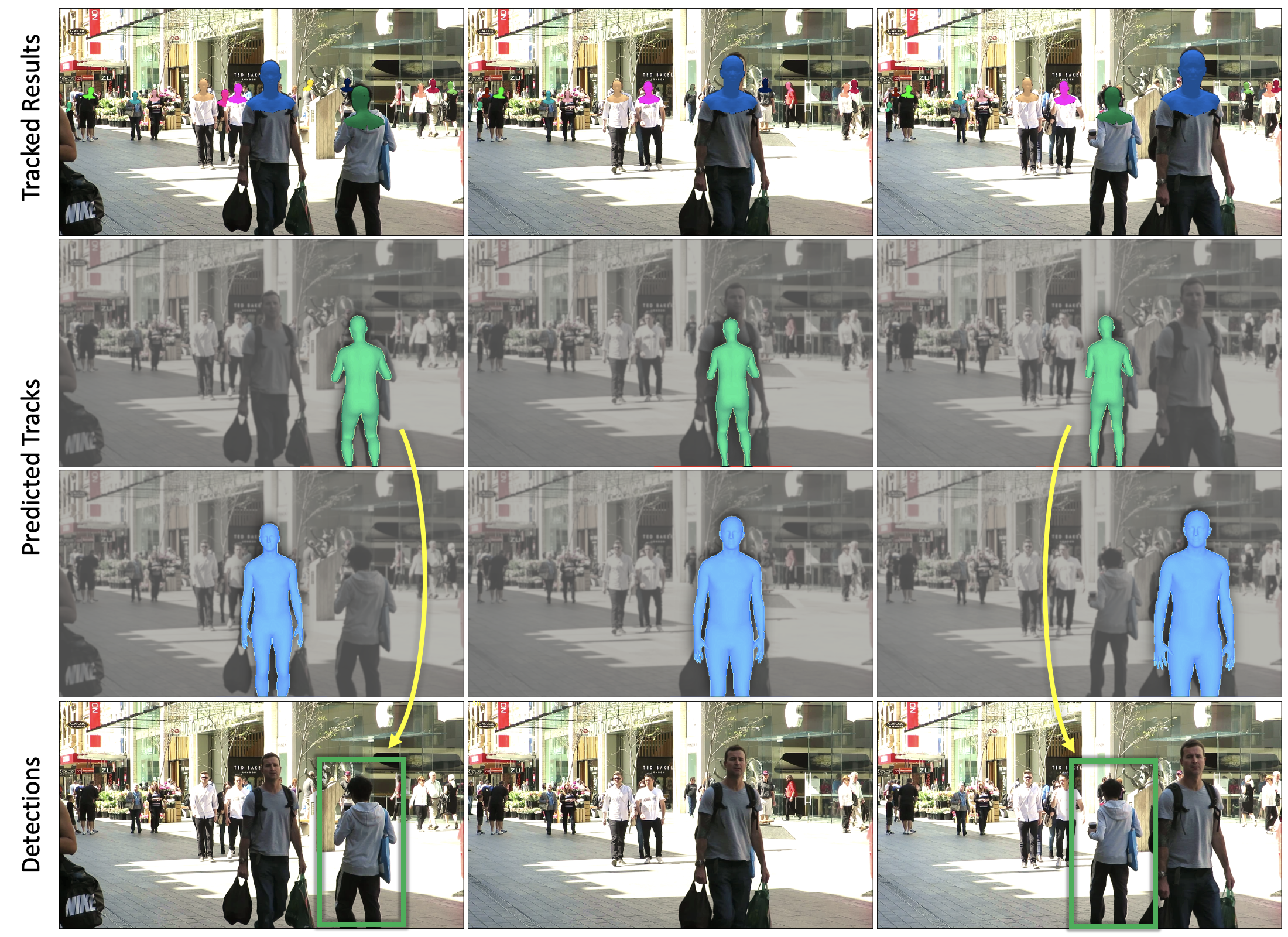}
    \caption{{\small \bf Tracking people by predicting and matching in 3D.} The top row shows our tracking results at three different frames. The results are visualized by a colored head-mask for unique identities. The second and third rows show renderings of the 3D states of the two people in their associated tracklets. The bottom row shows the bottom-up detections in each image frame which, after being lifted to 3D, will be matched with the 3D predictions of each tracklet in the corresponding frame. Note how in the middle frame of second row, the 3D representation of the person persists even though he is occluded in the image. More videos at \href{https://brjathu.github.io/PHALP/}{project site}.  }\label{fig:teaser}
\end{figure}

When we watch a video, we can segment out individual people, cars, or other objects and track them over time. The corresponding task in computer vision has been studied for several decades now, with a fundamental choice being whether to do the tracking in 2D in the image plane, or of 3D objects in the world. The former seems simpler because it obviates the need for inferring 3D, but if we do take the step of back-projecting from the image to the world, other aspects such as dealing with occlusion become easier. In the 3D world the tracked object doesn't disappear, and even young infants are aware of its persistence behind the occluder. A recent paper, Rajasegaran~\etal~\cite{rajasegaran2021tracking} argues convincingly on  the 3D side of this debate for people tracking, and presents experimental evidence that indeed performance is better with 3D representations. In this paper, we will take this as granted, and proceed to develop a system in the 3D setting of the problem. While our approach broadly applies to any object category where parameterized 3D models are available and can be inferred from images, we will limit ourselves in this paper to studying people,  the most important case in practice.

\begin{figure*}[!t]
    \centering
    \includegraphics[trim={0 0 0 0.5cm},clip,width=1.\textwidth]{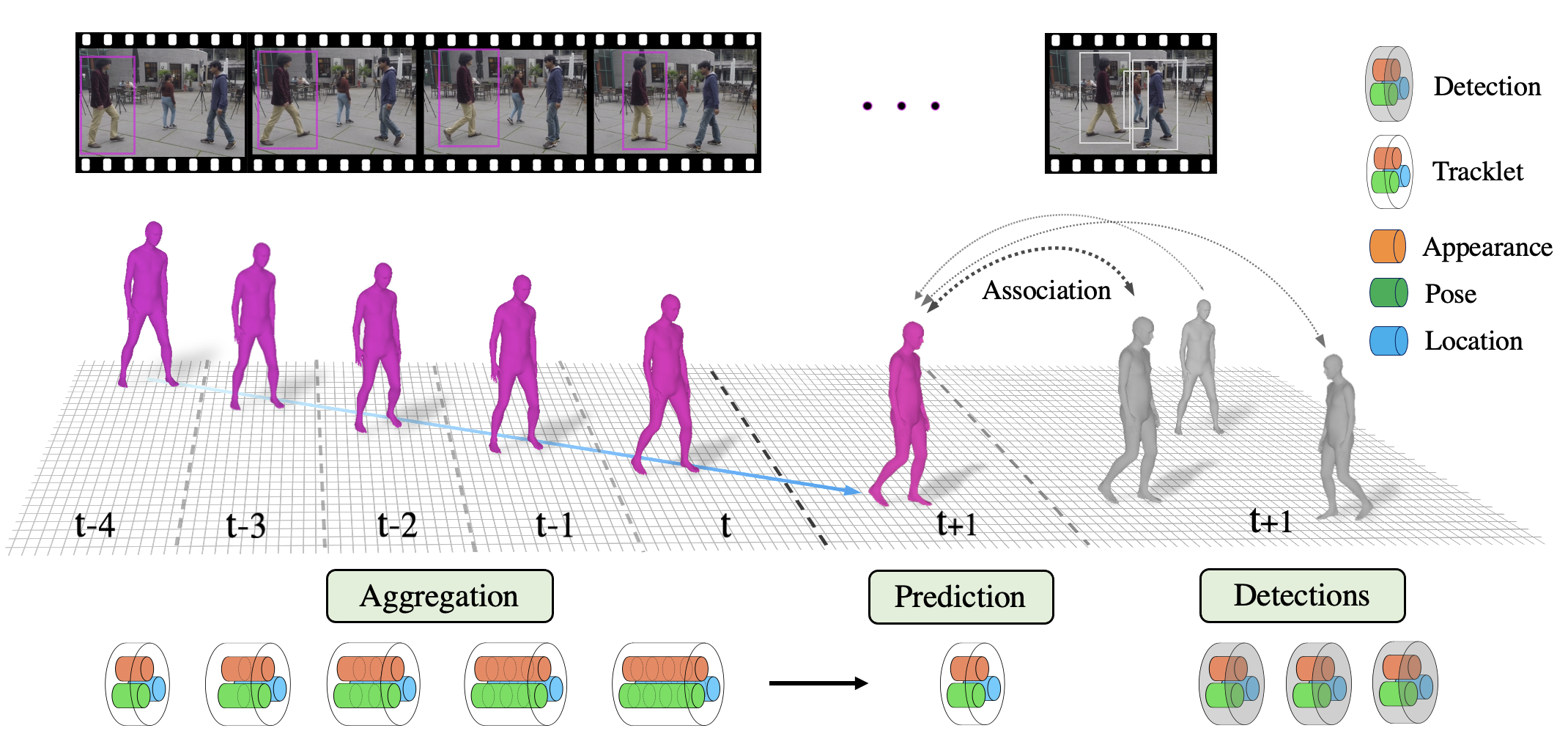}
    \caption{{\bf PHALP: Predicting Human Appearance, Location and Pose for Tracking}: We perform tracking of humans in 3D from monocular video. For every input bounding box, we estimate a 3D representation based on the 3D appearance, 3D pose and 3D location of the person. During tracking, these are integrated to form corresponding tracklet-based representations. We perform tracking by predicting the future representation of each person and using it to solve for association given the detected bounding boxes of a future frame.}\label{fig:main}
\end{figure*}

Once we have accepted the philosophy that we are tracking 3D objects in a 3D world, but from 2D images as raw data, it is natural to  adopt the vocabulary from control theory and estimation theory going back to the 1960s. We are interested in the ``state'' of objects in 3D, but all we have access to are ``observations'' which are RGB pixels in 2D. In an online setting, we observe a person across multiple time frames, and keep recursively updating our estimate of the person's state --- his or her appearance, location in the world, and pose (configuration of joint angles). Since we have a dynamic model (a ``tracklet''), we can also predict states at future times. When the next image frame comes in, we detect the people in it, lift them to 3D, and in that setting solve the association problem between these bottom-up detections and the top-down predictions of the different tracklets for this frame. Once the observations have been associated with the tracklets, the state of each person is re-estimated and the process continues. Fig.~\ref{fig:teaser} shows this process at work on a real video. Note that during a period of occlusion of a tracklet, while no new observations are coming in, the state of the person keeps evolving following his or her dynamics. It is not the case that ``Out of sight, out of mind''!

In an abstract form, the procedure sketched in the previous paragraph is basically the same as that followed in multiple computer vision papers from the 1980s and 1990s. The difference is that in 2021 we can actually make it work thanks to the advances brought about by deep learning and big data, that enable consistent and reliable lifting of people to 3D. For this initial lifting, we rely on the HMAR model~\cite{rajasegaran2021tracking}. This is applied on every detected bounding box of the input video and provides us with their initial, single frame, observations for 3D pose, appearance as well as location of the person in the 3D space.

As we link individual detections into tracklets, these representations are aggregated across each tracklet, allowing us to form temporal models, \ie, functions for the aggregation and prediction of each representation separately (see left side of Fig.~\ref{fig:main}). More specifically, for appearance, we use the canonical UV map of the SMPL model to aggregate appearance, and employ its most recent version as a prediction of a person's appearance. For pose, we aggregate information using a modification of the HMMR model~\cite{kanazawa2019learning}, where through its ``movie strip'' representation, we can produce 3D pose predictions. Finally, for 3D location, we use linear regression to predict the future location of the person.

This modeling enables us to develop our tracking system, PHALP (Predicting Human Appearance, Location and Pose for tracking), which aggregates information over time, uses it to predict future states, and then associates the predictions with the detections. First, we predict the 3D location, 3D pose and 3D appearance for each tracklet for a short period of time (right side of Fig.~\ref{fig:main}). For a future frame, these predictions need to be associated with the detected people of the frame. To measure similarity, we adopt a probabilistic interpretation and compute the posterior probabilities of every detection belonging to each one of the tracklets, based on the three basic attributes. With the appropriate similarity metric, association is then easily resolved by means of the Hungarian algorithm. The newly linked detections can now update the temporal model of the corresponding tracklets for 3D pose, 3D appearance and 3D location in an online manner and we continue the procedure by rolling-out further prediction steps. The final output is an identity label for each detected bounding box in the video. Notably, this approach can also be applied on videos with shot changes, e.g., movies~\cite{gu2018ava}, with minor modifications. Effectively, we only modify our similarity to include only appearance and 3D pose information for these transitions, since they (unlike location) are not affected by the shot boundary.

%% file: 03_related_work.tex
\section{Related work}
\label{sec:related}

\paragraph{Tracking.}
Object tracking is studied in various settings such as single object tracking, multi-object tracking for humans, and multi-object tracking for vehicles etc. The tracking literature is vast and we refer readers to~\cite{dendorfer2021motchallenge, yilmaz2006object, ciaparrone2020deep} for a comprehensive summary. In general tracking can be designed for any generic category, however, in this section we discuss the methods that focus on tracking humans. These approaches mostly work in a tracking by detection setting, where 2D location, key-point features~\cite{girdhar2018detect,snower202015,xiao2018simple} and 2D appearance~\cite{bergmann2019tracking,meinhardt2021trackformer,xu2019spatial,xu2020train} is used to associate detections over time. Quality of the detection plays a key role in tracking-by-detection setting and many works jointly learn or fine-tune their own detection models~\cite{bergmann2019tracking,meinhardt2021trackformer}. In this work, we are interested in the effectiveness of 3D representations for tracking and thus assume that detection bounding boxes are provided, which we associate through our representations. On the other hand, tracking by regression predicts future locations using the knowledge of the past detections. While this alleviate the requirement for good quality detections, most of the works regress in the image plane. The projection from 3D world to the image plane makes it hard to make this prediction, therefore these methods need to learn non-linear motion models~\cite{alahi2016social, zhang2020long, bergmann2019tracking}. Compared to these methods, PHALP predicts short-term location in 3D coordinates, by simple linear regression. Additionally, we also predict appearance and pose features for better association. 

Finally, there are methods that incorporate 3D information in tracking, however these approaches assume multiple input cameras~\cite{kwon2020recursive, zhang20204d} or 3D point cloud observation from lidar data~\cite{weng2020gnn3dmot}. In this paper we focus on the setting where the input is a monocular video. Some recent works tracks occluded people based on the object permanence~\
\cite{khurana2021detecting, tokmakov2021learning}. These methods learn complex functions to predict the locations of occluded people. However, by placing humans in 3D space and predicting their location, pose and appearance, object permanence is already built into our system. 

\paragraph{Monocular 3D human reconstruction.}
Although there is a long history of methods for 3D human reconstruction from monocular images, \eg,~\cite{bregler1998tracking,guan2009estimating}, here we focus on more recent works. Many of the relevant approaches rely on the SMPL model~\cite{loper2015smpl}, which offers a low dimensional parameterization of the human body. HMR~\cite{kanazawa2018end} has been one of the most notable ones, using a neural network to regress the parameters of a SMPL body from a single image. Follow-up works have improved the robustness of the original model~\cite{kocabas2021pare, kolotouros2019learning}, or added additional features like estimation of camera parameters~\cite{kocabas2021spec}, or probabilistic estimation of pose~\cite{kolotouros2021probabilistic}. Recently, Rajasegaran~\etal~\cite{rajasegaran2021tracking} introduced HMAR, by extending the model with an appearance head. Other works have focused on extending HMR to the temporal dimension,~\eg, HMMR~\cite{kanazawa2019learning}, VIBE~\cite{kocabas2020vibe}, MEVA~\cite{luo20203d} and more~\cite{pavlakos2020human}. In this work, we make use of a modification of the HMAR model~\cite{rajasegaran2021tracking} as the main feature backbone, while also employing a model that follows the HMMR principles~\cite{kanazawa2019learning} for temporal pose prediction, but instead, using a transformer~\cite{vaswani2017attention} to aggregate pose information over time. Regarding human motion prediction, Kanazawa~\etal~\cite{kanazawa2019learning}, regress future poses from the temporal pose representation of HMMR, the ``movie-strip''. Zhang~\etal~\cite{zhang2019predicting} extend this to PHD, employing autoregressive prediction of human motion. Aksan~\etal~\cite{aksan2020spatio} also regress future human motion in an autoregressive manner, using a transformer.

%% file: 04_technical_approach.tex
\section{Method}
\label{sec:technical}

Tracking humans using 3D representations has significant advantages, including that appearance is independent of pose variations and the ability to have amodal completion for humans during partial occlusion. Our tracking algorithm accumulates these 3D representations over time, to achieve better association with the detections. \ours has three main stages: 1) lifting humans into 3D representations in each frame, 2) aggregating single frame representations over time and predicting future representations, 3) associating tracks with detections using predicted representations in a probabilistic framework. We explain each stage in the next sections.

\subsection{Single-frame processing} 
The input to our system is a set of person detections along with their estimated segmentation masks, provided by conventional detection networks, like Mask-RCNN~\cite{he2017mask}. Each detection is processed by our feature extraction backbone that computes the basic representations for pose, appearance and location on a single-frame basis. For this feature extraction we use a modification of the HMAR model~\cite{rajasegaran2021tracking}. HMAR returns a feature representation for the 3D pose $\mathbf{p}$, for appearance $\mathbf{a}$, while it can recover an estimate for the 3D location $\mathbf{l}$ for the person.

The standard HMAR model takes as input the pixels in the bounding box corresponding to a detected person. This means that in a crowded, multi-person scenario, the input will contain pixels corresponding to more than one person in the bounding box, potentially confusing the network. To deal with this problem, we modify HMAR to take as additional input,  the pixel level mask of the person of interest (this is readily available as part of the output of Mask R-CNN) and re-train HMAR.  Obviously, we cannot expect this step to be perfect, since there can be inaccuracies in the bounding box detections or mask segmentations. However, we observed that the model gives more robust results in the case of close person-person interactions, which are common in natural videos.

\subsection{3D tracklet prediction}
\label{sec:tracklet}

The 3D estimates for each detection provide a rich and expressive representation for each bounding box. However, they are only the result of single-frame processing. During tracking, as we expand each tracklet, we have access to more information that is representative of the state of the tracklet along the whole trajectory. To properly leverage this information, our tracking algorithm builds a tracklet representation during every step of its online processing, which allows us to also predict the future states for each tracklet. In this section we describe how we build this tracklet representation, and more importantly, how we use it to {\it predict} the future state of each tracklet.

\begin{figure}[t]
     \centering
     \begin{subfigure}{0.48\textwidth}
         \centering
         \includegraphics[width=1\textwidth]{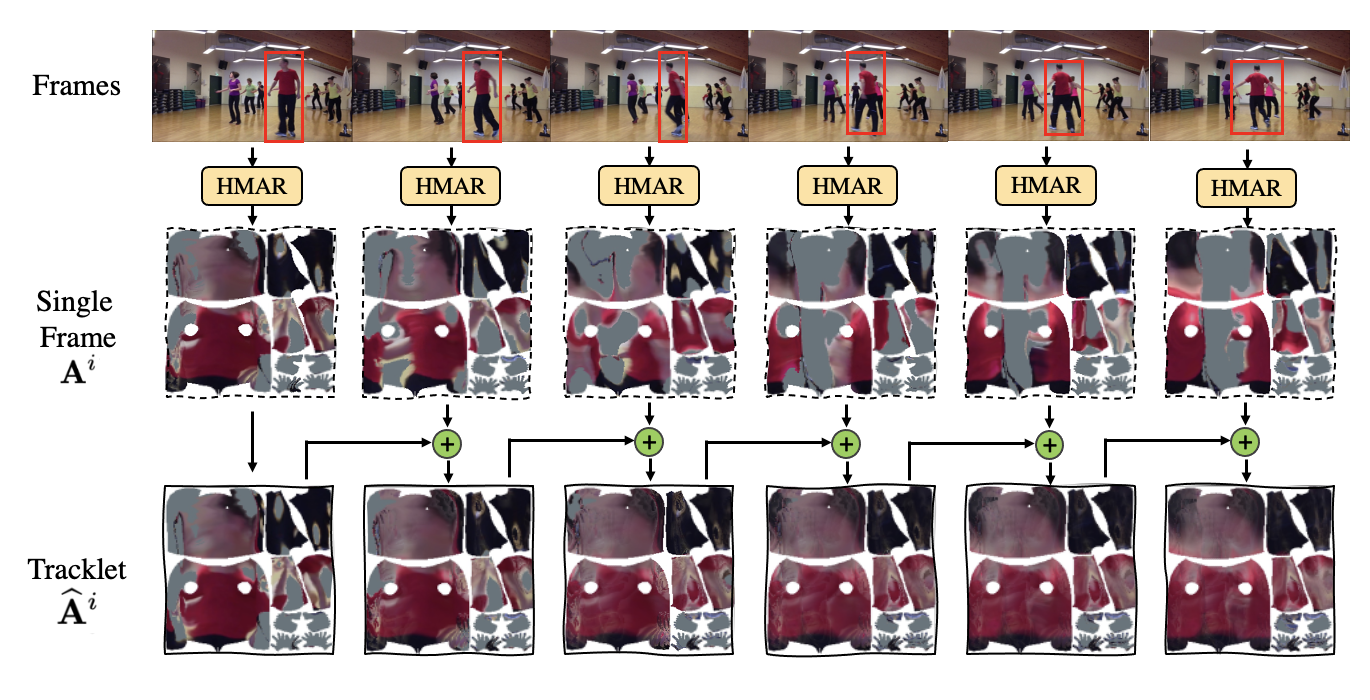}
     \end{subfigure} 
     \caption{\textbf{Prediction of appearance:} We show how the single frame appearance $\mathbf{A}^i$ is aggregated over time for the prediction of the tracklet appearance $\widehat{\mathbf{A}}^i$. At the start, we only see the front side of the {\it person indicated} in the frame, however as he moves his visibility changes, and we only see his back side. With the single frame appearance, we can see that the visibility changes corresponding to the visibility of the person in the frame. However, in the tracklet, the appearance is accumulated over time, and even if the front side is not visible in the last frame, we can see that the tracklet has predicted these regions using the past frames.}
     \label{fig:appearance_aggregation}
\end{figure}

\vspace{0.2cm}
\noindent{\bf Appearance:}
The appearance pathway is used to integrate appearance information for each person over multiple frames. The single frame appearance representation for the person $i$ at time step $t$, $\mathbf{A}^{i}_t$, is taken from the HMAR model by combining the UV image of that person $\mathbf{T}^{i}_t \in \mathcal{R}^{3 \times 256 \times 256}$ and the corresponding visibility map $\mathbf{V}^{i}_t \in \mathcal{R}^{1 \times 256 \times 256}$ at time step $t$: 
\begin{align*}
    \mathbf{A}^{i}_t = [\mathbf{T}^{i}_t, \mathbf{V}^{i}_t] \in \mathcal{R}^{4 \times 256 \times 256}
\end{align*}
Note that the visibility mask $\mathbf{V}^{i}_t \in [0,1]$ indicates whether a pixel in the UV image is visible or not, based on the estimated mask from Mask-RCNN. Now, if we assume that we have established the identity of this person in neighboring frames, we can integrate the partial appearance information coming from the independent frames to an overall tracklet appearance for the person. Using the set of single frame appearance representations $\mathcal{A}^i=\{\mathbf{A}^{i}_{t}, \mathbf{A}^{i}_{t-1}, \mathbf{A}^{i}_{t-2}, ... \}$, after every new detection we create a singe per-tracklet appearance representation:
\begin{align*}
    \widehat{\mathbf{A}}^{i}_t &= \Phi_A(\widehat{\mathbf{A}}^{i}_{t-1},  \mathbf{A}^{i}_t) = (1-\alpha) * \widehat{\mathbf{A}}^{i}_{t-1} + \alpha \mathbf{A}^{i}_t \\
    \text{where, }  \ \ \ \alpha &= \begin{cases}
      \alpha_0, & \text{if}\  \ \ \widehat{\mathbf{V}}^{i}_{t-1}=1 \ \ \text{and} \ \ \mathbf{V}^{i}_t=1 \\
      1, & \text{if}   \ \ \ \widehat{\mathbf{V}}^{i}_{t-1}=0 \ \ \text{and} \ \ \mathbf{V}^{i}_t=1 \\
      0, & \text{if} \ \ \ \widehat{\mathbf{V}}^{i}_{t-1}=1 \ \ \text{and} \ \ \mathbf{V}^{i}_t=0.
    \end{cases}
\end{align*}
Here, $\Phi_A$ is the appearance aggregation function, 
which takes a weighted sum of the previous tracklet appearance representation and the new detection appearance representation. Note that, at the start of the tracklet we simply assign the initial single-frame representation to the tracklet representation ($\widehat{\mathbf{A}}^{i}_0 = \mathbf{A}^{i}_0$). With this definition of $\Phi_A$, we can aggregate appearance information over time, while allowing the representation to change slowly to account for slight appearance changes of the person during a video. Moreover, the UV image provides appearance of each point on the body surface independently of body pose and shape which enables the simple summation operation on the pixel space, without any learnable components. Figure~\ref{fig:appearance_aggregation} shows how the UV image of the person is aggregated over time and used for association of new detections. 

For appearance prediction, we make the realistic assumption that human appearance will not change rapidly over time. Then, the appearance of the tracklet $\widehat{\mathbf{A}}^{i}_t$ can function as a reasonable prediction for the future appearance of the person. Therefore, we use $\widehat{\mathbf{A}}^{i}_t$ as the prediction for appearance and use it to measure similarity against a detection in the future frames.

\vspace{0.2cm}
\noindent{\bf Location:}
Lifting humans from pixels into the 3D space allows us to place them in the global 3D location. Let us assume that a person $i$ at time $t$ has an estimated 3D location $\mathbf{l}^{i}_t$. Although, we can get an estimate for the location of the person in the global camera frame, this tends to be noisy, particularly along the $z$-axis. To avoid any instabilities when it comes to predicting future location, instead of performing our prediction on the Euclidean $(X,Y,Z)^T$ space, we express our locations in an equivalent $\mathbf{l}^{i}_t = (x,y,n)^T$ space where $(x,y)$ is the location of the root of the person in the pixel space and $n$ is {\it nearness}, defined as log  inverse depth $n = \log(1/z)$. Nearness is a natural parameterization of depth in multiple vision settings, \eg,~\cite{koenderink1986optic}, because of the $1/z$ scaling of perspective projection. In our case it corresponds to the scale of the human figures that we estimate directly from images. We independently linearly regress the location predictions for $x, y$ and $n$. This is somewhat like the Constant Velocity Assumption (CVA) used in past tracking literature, but there is a subtlety here because constant velocity in 3D need not give rise to constant velocity in 2D (a person would appear to speed up as she approaches the camera). But local linearization is always a reasonable approximation to make, which is what we do.

 Let us assume that a tracklet has a set of past locations $\mathcal{L}^{i} = \{\mathbf{l}^{i}_t, \mathbf{l}^{i}_{t-1}, \mathbf{l}^{i}_{t-2}, ...\}$. Then, the prediction of the location for time step $t+1$ is given by:
\begin{align*}
    \widehat{\mathbf{l}}^{i}_{t+1} &= (\widehat{x}^{i}_{t+1},\widehat{y}^{i}_{t+1},\widehat{n}^{i}_{t+1})^T \\
   \ \text{where,}  \ \ \widehat{x}^{i}_{t+1} &= \Phi_L(\{x^{i}_{t}, x^{i}_{t-1}, x^{i}_{t-2}, ..., x^{i}_{t-w}\}, t+1).
\end{align*}
Here, $\Phi_L$ is the location aggregation function and we use a simple linear regression for prediction in our tracking algorithm. $\widehat{y}^{i}_{t+1}$ and $\widehat{n}^{i}_{t+1}$ are also predicted in a similar fashion. $\Phi_L$ takes the last $w$ observations to fit a line by least squares and regress the future location for $x, y$ and $n$ independently. From the standard theory of linear regression, the prediction interval for $x$ at a time step $t'$ is given by the equation below:
\begin{align*}
\tiny
    \delta_x(t') = t_{(1-\alpha/2)} \times \sqrt{MSE \times \left( 1 + \frac{1}{w} + \frac{(t'-\bar{t})^2}{\sum(t-\bar{t})^2}\right)}.
\end{align*}
Here, $t_{(1-\alpha/2)}$ is the Student's $t$ distribution with confidence $\alpha$ and degree of freedom $w-2$. $MSE$ is the mean squared error on the predicted locations and $\bar{t}$ is the mean of the time stamps for the previous observations. In a similar manner, we can compute prediction intervals $\Delta_y, \Delta_n$ for $y$ and $n$ respectively.

\vspace{0.2cm}
\noindent{\bf Pose:}
For the pose pathway, we need to integrate pose information across the tracklet and be able to predict future poses for the near future. To do this, we borrow ideas from the HMMR architecture~\cite{kanazawa2019learning}. Effectively, we learn a function $\Phi_P$ that takes as input a series of pose embeddings of a person $\mathcal{P}^i = \{\mathbf{p}^i_t, \mathbf{p}^i_{t-1}, \mathbf{p}^i_{t-2}, ...\}$ and computes a temporal pose embedding $\widehat{\mathbf{p}}_t$. We train this temporal pose aggregation function $\Phi_P$ to smooth the pose $\widehat{\mathbf{p}}^i_{t}$ at frame $t$, and regress the future pose representations $\{\widehat{\mathbf{p}}^i_{t+1}, \widehat{\mathbf{p}}^i_{t+2}, ..., \widehat{\mathbf{p}}^i_{t+c} \}$ (typically for up to $c=12$ frames in the future).  We use a transformer~\cite{vaswani2017attention} to compute $\Phi_P$. This choice allows for some additional flexibility, since sometimes we are not able to detect an identity in some frames (\eg, due to occlusions), which can be handled gracefully by the transformer, by masking out the attention for the representation of the corresponding frame.

\begin{figure}[t]
     \centering
     \begin{subfigure}{0.48\textwidth}
         \centering
         \includegraphics[width=1\textwidth]{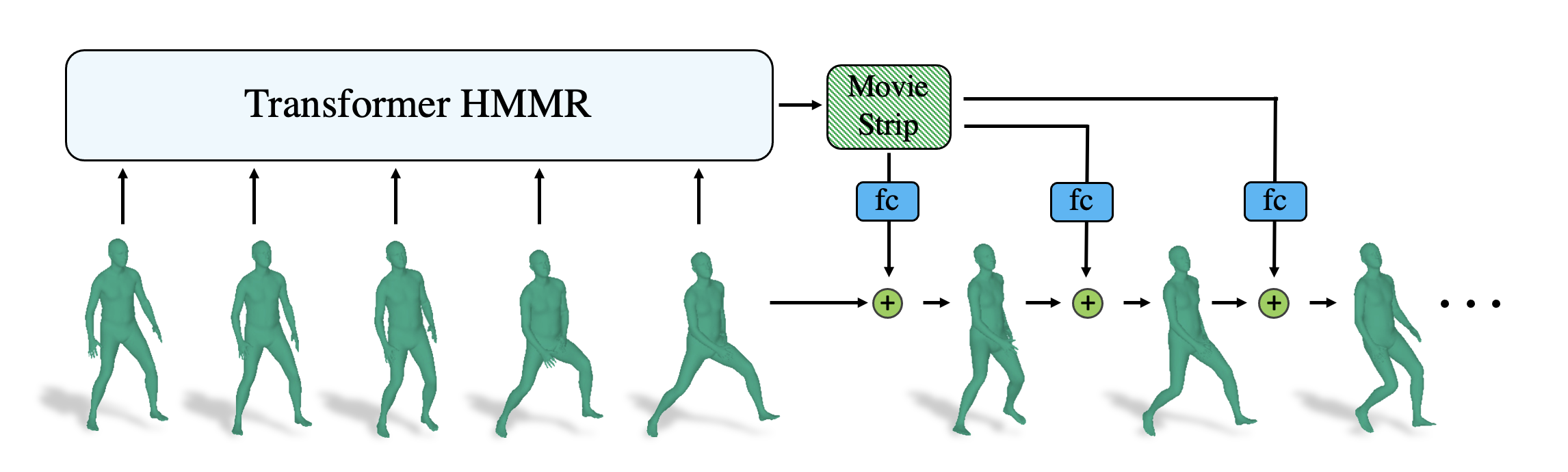}
     \end{subfigure}
     \caption{\textbf{Prediction of Pose:} We use a modified version of HMMR~\cite{kanazawa2019learning} with transformer backbone. Having transformer as the backbone gives us the flexibility to have missing people in the tracklet (by masking the attention maps), while still allowing us to predictions of future poses. Finally, the transformer give us a movie-strip representation and that is used to regress future poses. }
     \label{fig:pose_aggregation}
\end{figure}

\subsection{Tracking with predicted 3D representations}
\label{sec:tracking}

\begin{figure}[t]
     \centering
     \begin{subfigure}{0.22\textwidth}
         \centering
         \includegraphics[width=1.1\textwidth]{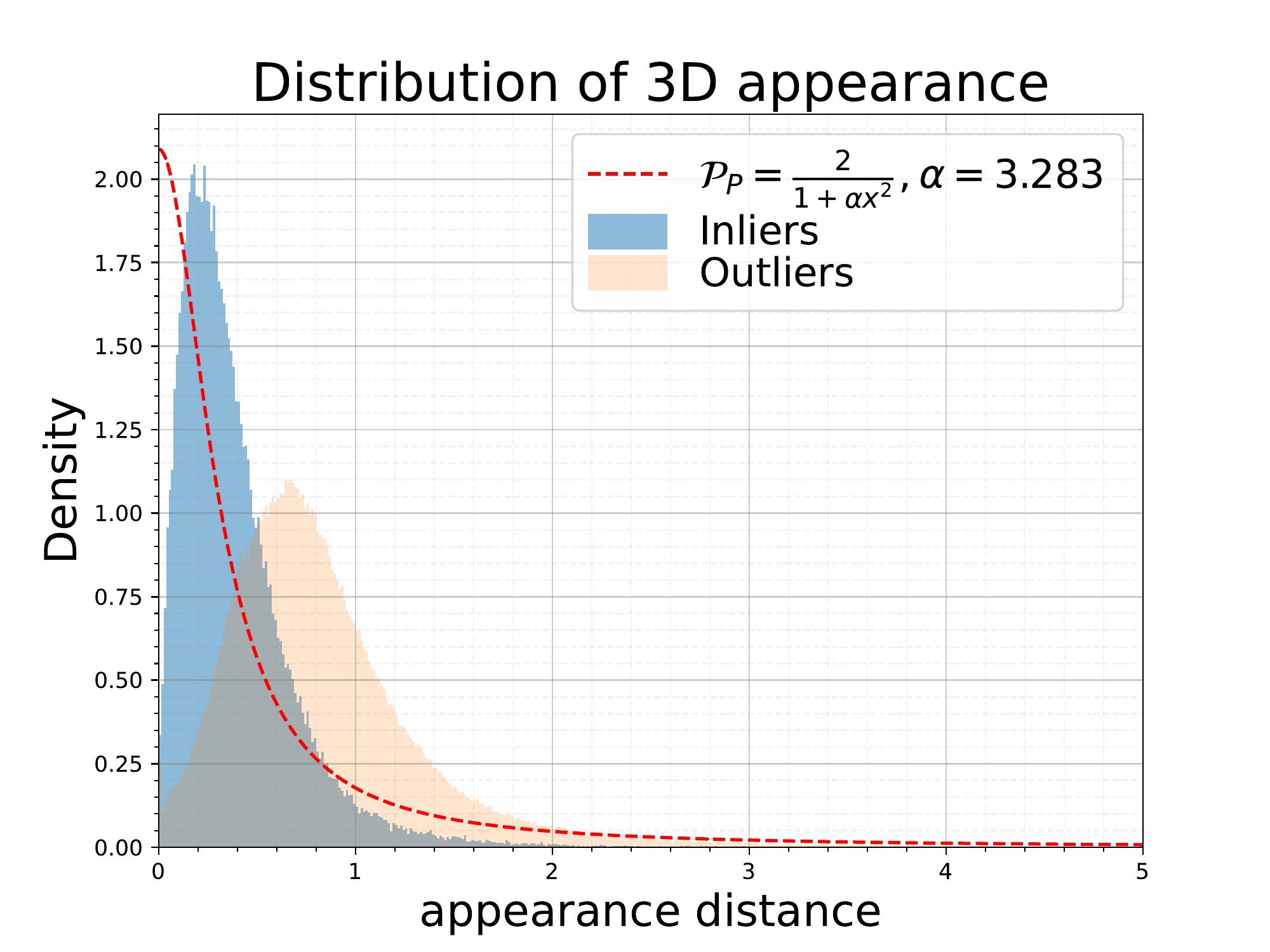}
     \end{subfigure}
     \begin{subfigure}{0.22\textwidth}
         \centering
         \includegraphics[width=1.1\textwidth]{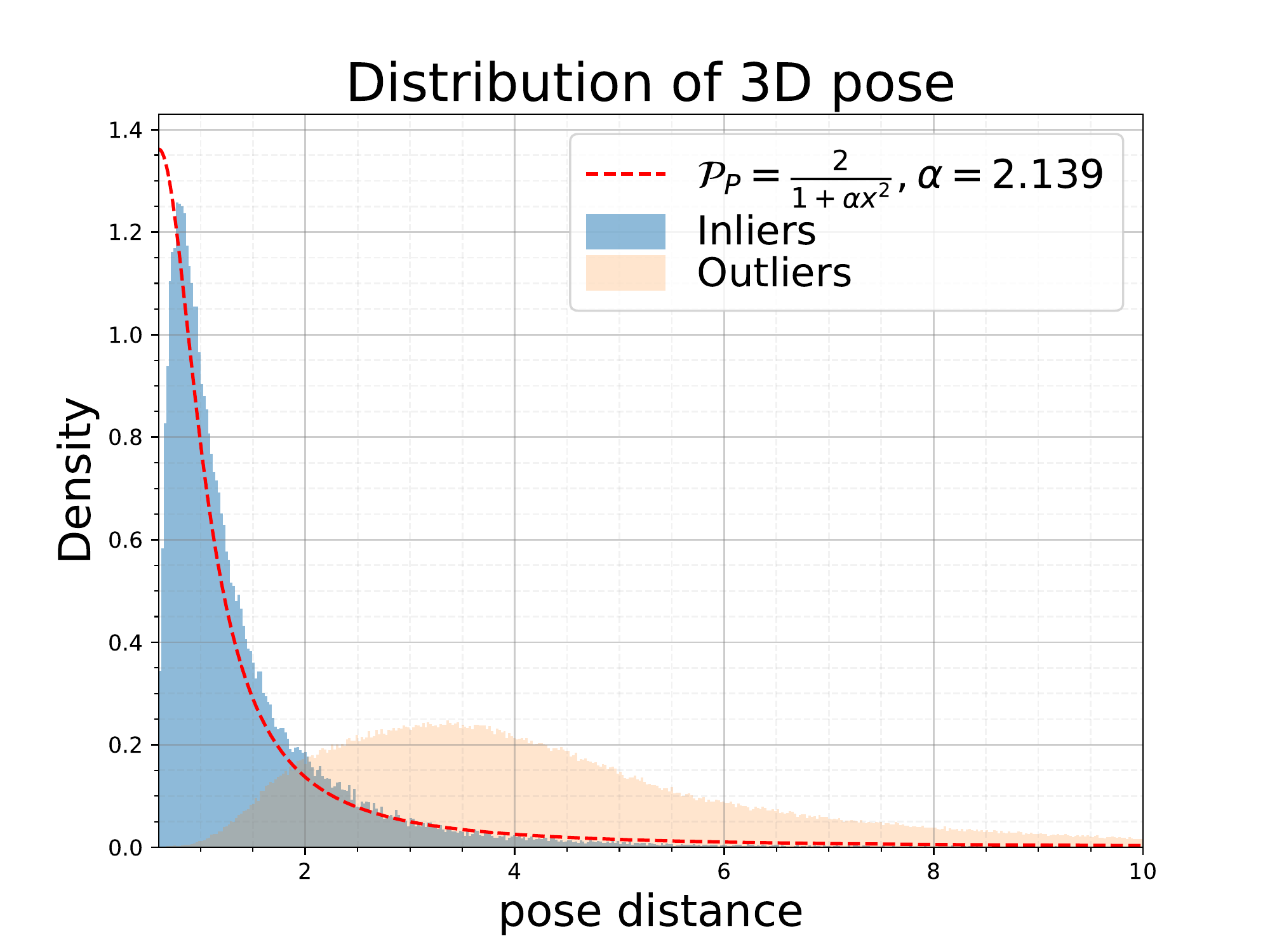}
     \end{subfigure}  \\
     \begin{subfigure}{0.22\textwidth}
         \centering
         \includegraphics[width=1.1\textwidth]{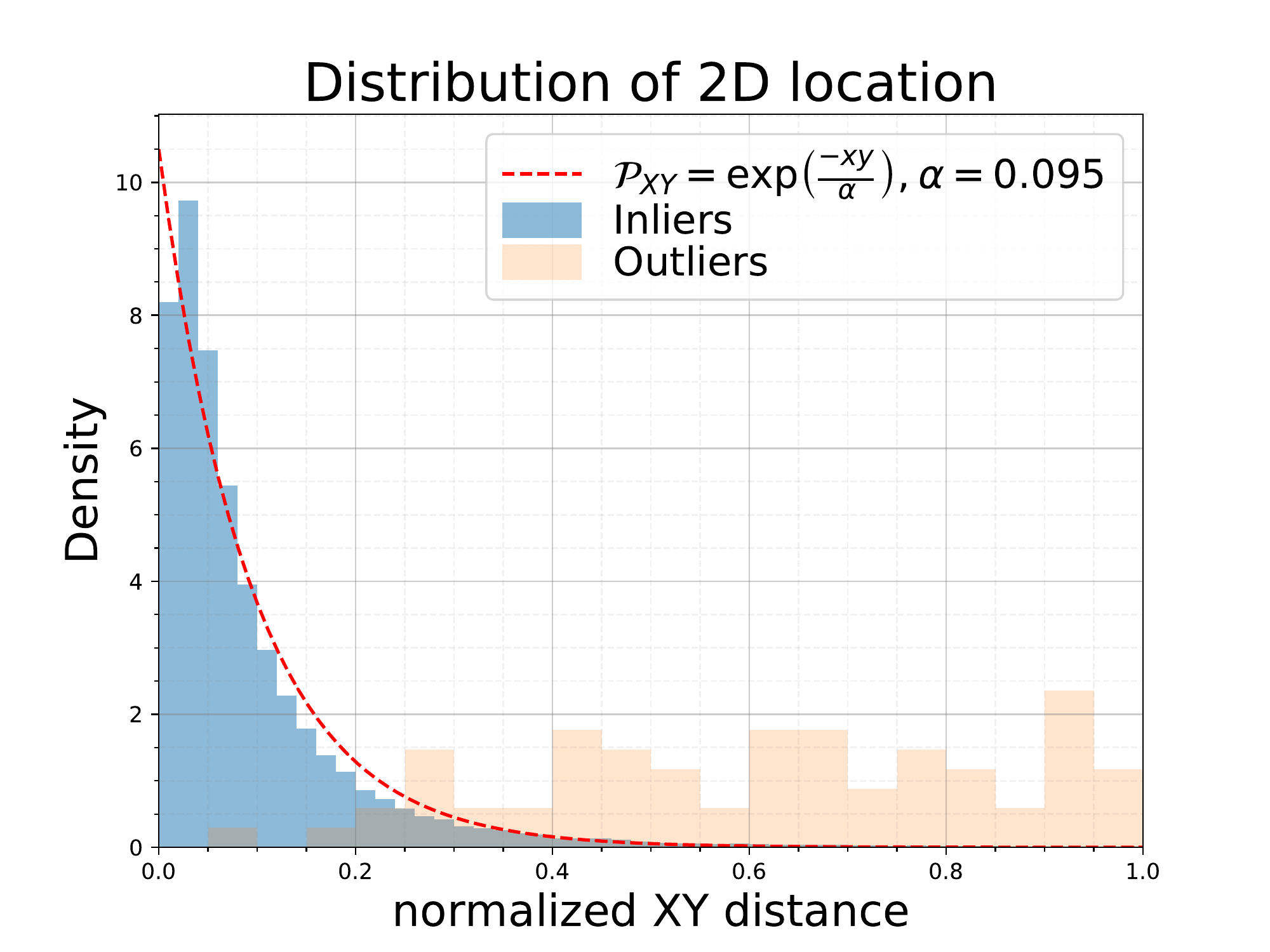}
     \end{subfigure}
     \begin{subfigure}{0.22\textwidth}
         \centering
         \includegraphics[width=1.1\textwidth]{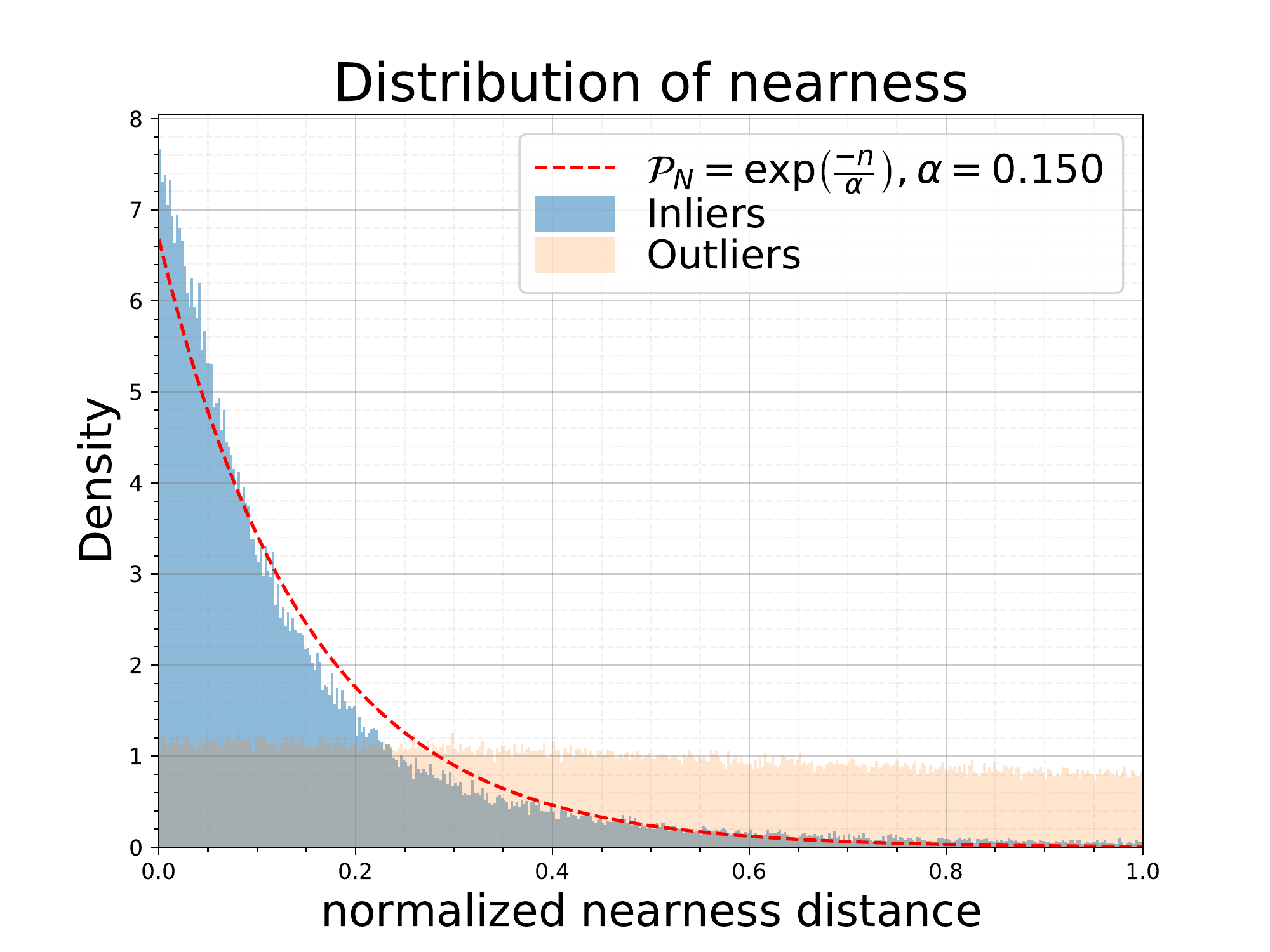}
     \end{subfigure}
     \caption{{ \textbf{Conditional distributions of the attribute distances:} We plot the data for the distances between the tracklet prediction and the single frame detection using the ground truth data from PoseTrack~\cite{andriluka2018posetrack}. The curves show how the correct matches (inliers) and the incorrect matches (outliers) are distributed. Note that, for 2D location and nearness we plot the distances normalized by the prediction interval. }}
     \label{fig:distributions}
     \vspace{-0.4cm}
\end{figure}

Given the bounding boxes and their single-frame 3D representations, our tracking algorithm associates identities across frames in an online manner. At every frame, we make future predictions for each tracklet and we measure the similarity with the detected single-frame representation. More specifically, let us assume that we have a tracklet $T_i$, which has been tracked for a sequence of frames and has information for appearance, pose and location. The tracklet predicts its appearance $\widehat{\mathbf{A}}$, location $\widehat{\mathbf{l}}$ and pose $\widehat{\mathbf{p}}$ for the next frame, and we need to measure a similarity score between these predictions of the track $T_i$ and a detection $D_j$ to make an association. Our tracklet representation has three different attributes (appearance, location and pose), so, directly combining their similarities/distances would not be ideal, since, each attribute has different characteristics. Instead, we investigate the conditional distributions of inliers and outliers of the attributes. Figure~\ref{fig:distributions} presents the corresponding probability distributions for the PoseTrack dataset~\cite{andriluka2018posetrack}. The characteristics of these distributions motivate our design decisions for our further modeling. 

Assume that tracklet $T_i$ has an appearance representation $\widehat{\mathbf{A}}^i_{t}$. On the detection side, the detection $D_j$ has a single-frame appearance representation $\mathbf{A}^j_t$. Both of these representations are in the pixel space, therefore we first encode them into an embedding space using the HMAR appearance-encoder network. This gives us an appearance embedding $\widehat{\mathbf{a}}^i_t$ and $\mathbf{a}^j_t$ for the prediction of the tracklet $T_i$ and detection $D_j$, respectively. We are interested in estimating the posterior probability of the event where the detection $D_j$ belongs to the track $T_i$, given some distance measure of the appearance feature ($\Delta_a$). Assuming that the appearance distance is $\Delta_a = ||\widehat{\mathbf{a}}^i_t - \mathbf{a}^j_t||_2^2$, then the posterior probability is proportional to the conditional probability of the appearance distances, given correct assignments based on Bayes rule. We model this conditional probability as a Cauchy distribution, based on the observations from the inlier distribution of appearance distances (see also Fig~\ref{fig:distributions}):
\begin{align*}
    \mathcal{P}_A(D_j \in T_i| d_a = \Delta_a) \propto \frac{1}{1+\beta_a\Delta_a}
\end{align*}
The distribution has one scaling hyper-parameter $\beta_a$. 

Similarly, for pose, we use Cauchy distribution to model the conditional probability of inlier distances.  We measure pose distance $\Delta_p = ||\widehat{\mathbf{p}}^i_t - \mathbf{p}^j_t||_2^2$ between the predicted pose representation $\mathbf{p}^i_t$ from the track $T_i$ and the pose representation $\mathbf{p}^j_t$ of detection $D_j$. The posterior probability that the detection belongs to the track, given the pose distance is:
\begin{align*}
    \mathcal{P}_P(D_j \in T_i| d_p = \Delta_p) \propto \frac{1}{1+\beta_p\Delta_p}
\end{align*}
Here, $\Delta_p=||\widehat{\mathbf{p}}^i_t - \mathbf{p}^j_t||_2^2$ and $\beta_p$ is the scaling factor. 

For location, let us assume the track $T_i$ has predicted a location $\widehat{\mathbf{l}}^i_t=(\widehat{x}^i_t, \widehat{y}^i_t, \widehat{n}^i_t)^T$ with a set of prediction intervals $\{\delta_x, \delta_y, \delta_n \}$, and the detection $D_j$ is at a 3D location $\mathbf{l}^j_t=(x^j_t, y^j_t, n^j_t)^T$. We treat the 3D coordinates $x, y$ and the nearness term $n$ coordinates independently, and compute the posterior probabilities of the detection belongs to the tracklet given the location distance. We model the conditional probability distribution as an exponential distribution, based on the findings from the empirical data. The Fig~\ref{fig:distributions} shows the distribution of 2D distance and nearness distance, scaled by the confidence interval, of inliers approximately follow the exponential distribution.  
\begin{align*}
    \mathcal{P}_{XY}(D_j \in T_i| d_{xy}=\Delta_{xy}) \propto \frac{1}{\beta_{xy}}\exp\left({\frac{-\Delta_{xy}}{\beta_{xy} \delta_{xy}}}\right)
\end{align*}
Here, $\beta_{xy}$ is a scaling parameter for the exponential distribution, $\Delta_{xy}$ is the 2D pixel distance between the predicted track and the detection and $\delta_{xy} = \sqrt{\delta_x^2 + \delta_y^2}$ is the prediction interval for the 2D location prediction. We use a similar form of exponential distribution for the posterior probability for nearness $\mathcal{P}_{N}$: 
\begin{align*}
    \mathcal{P}_{N}(D_j \in T_i| d_{n}=\Delta_{n}) \propto \frac{1}{\beta_{n}}\exp\left({\frac{-\Delta_{n}}{\beta_{n} \delta_{n}}}\right)
\end{align*}
Here, $\beta_n$ is the scaling parameter for the exponential distribution, $\delta_{n}$ is the confidence interval for the nearness prediction, and $\Delta_{n}$ is the $L_1$ distance between the nearness of the tracklet prediction and the detection.

Now that we have computed the conditional probabilities of the detection belonging to a track conditioned on the individual cues of appearance, location and pose, we can compute the overall conditional probability of the detection $D_j$ belonging to the track $T_i$, given all the cues together (assumed to be independent):
\begin{align*}
    \mathcal{P}(D_j \in T_i| \Delta_a, \Delta_p, \Delta_{xy}, \Delta_n) \propto  \mathcal{P}_A \mathcal{P}_P\mathcal{P}_{XY} \mathcal{P}_N 
\end{align*}
This allow us to estimate how probable an association is based on various attribute distances. Finally, we map the similarity measures (probability values up to a scale), to cost values, for solving association. The cost function between the detection representations and a predicted representations of the tracklet is defined as:
\begin{align*} 
    \Phi_C(D_j, T_i) &= -\log(\mathcal{P}(D_j \in T_i)) \\
                     &= -\log(\mathcal{P}_A) - \log(\mathcal{P}_P) -\log(\mathcal{P}_{XY}) -\log(\mathcal{P}_N),
\end{align*}
where the second equality is up to an additive constant.
Once the cost between all the tracks and the detection is computed, we simply pass it to the Hungarian algorithm for solving the association.

\begin{algorithm}[!t]
\caption{Tracking Algorithm}
\label{alg:tracking}
\begin{algorithmic}[1]
\Procedure{PHALP Tracking}{}\\
\algorithmicrequire{ \ All active tracklets \ $\mathcal{T}$, all detections and their single frame 3D representations at time $t$, $\mathcal{D}$} and maximum age of a track $t_{max}$. 
\For {$T_j \in \mathcal{T}$} 
    \State {\color{PineGreen} \# predict all  attributes for the next frame.} 
    \State $\widehat{\mathbf{A}}^j_t \gets \Phi_A(\{\mathbf{A}^j_{t-1}, \mathbf{A}^j_{t-2},...\})$ 
    \State $\widehat{\mathbf{p}}^j_t \ \gets \Phi_P(\mathbf{p}^j_{t-1}, \mathbf{p}^j_{t-1},...\})$
    \State $\widehat{\mathbf{l}}^j_t \ \ \gets \Phi_L(\{\mathbf{l}^j_{t-1}, \mathbf{l}^j_{t-2},...\})$
\EndFor
\State {\color{PineGreen} \# Compute the cost between tracks and detections.} 
\State $\mathbf{C}_{i,j} \gets \Phi_C(D_i, T_j)$ for all $D_i \in \mathcal{D}$ and $T_j \in \mathcal{T}$ 
\State {\color{PineGreen} \# Hungarian to assign detections to tracklets.} 
\State $\mathcal{M}, \mathcal{T}_u, \mathcal{D}_u \gets \texttt{Assignment}(\mathbf{C})$
\State {\color{PineGreen} \# Update the matched tracks.} 
\State $\mathcal{T} \gets \{ T_j(D_i), \ \ \forall (i,j) \in \mathcal{M}\}$
\State {\color{PineGreen} \# Increase the age of unmatched tracks.} 
\State $\mathcal{T} \gets \{ T_j(age)+=1, \ \ \forall (j) \in \mathcal{T}_u\}$
\State {\color{PineGreen} \# Make new tracks with unmatched detections.} 
\State $\mathcal{T} \gets \{ T_j(D_i), \ \ \forall (i) \in \mathcal{D}_u,\ j=|\mathcal{T}+1| \}$
\State Kill the tracks with age $\geq$ $t_{max}$. \\
\Return Tracklets $\mathcal{T}$
\EndProcedure
\end{algorithmic}
\end{algorithm}

\vspace{0.2cm}
\noindent\textbf{Estimating the parameters of the cost function:} The cost function $\Phi_C$ has 4 parameters ($\beta_a, \beta_p, \beta_{xy} \ \text{and} \ \beta_n$). Additionally, the Hungarian algorithm has one parameter $\beta_{th}$ to decide whether the track is not a match to the detection. Therefore, overall we have five parameters for the whole association part of our tracking system. Now, we treat this as an empirical risk minimization problem and optimize the $\beta$ values based on a loss function. We initialize $\beta_a, \beta_p, \beta_{xy} \ \text{and} \ \beta_n$ with the values from the estimated density functions and use frame level association error as a loss function for the optimization. We use the Nelder–Mead~\cite{nelder1965simplex} algorithm for this optimization. Finally, the optimized $\beta$ values are used for the cost function across all the datasets, and a simple tracking algorithm is used to associate detections with tracklet predictions. The sketch of the tracking algorithm is shown in Algorithm~\ref{alg:tracking}.

\subsection{Extension to shot changes}
\label{sec:shots}

Our framework can easily be extended to also handle shot changes, which are common in edited media, like movies, TV shows, but also sports. Since shot changes can be detected relatively reliably, we use an external shot detector~\cite{huang2020movienet} to identify frames that indicate shot changes. Informed by the detection of this boundary, during tracking, we update the distance metric accordingly. More specifically, since appearance and 3D pose are invariant to the viewpoint, we keep these factors in the distance computation, while we drop the location distance from the distance metric, because of the change in the camera location. Then, the association is computed based on this updated metric. We use the AVA dataset~\cite{gu2018ava} to demonstrate this utility of our tracking system and present results in Section~\ref{sec:experiments}.

%% file: 05_experiments.tex
\section{Experiments}
\label{sec:experiments}

In this section, we present the experimental evaluation of our approach. We report results on three datasets: PoseTrack~\cite{andriluka2018posetrack}, MuPoTS~\cite{mehta2018single} and AVA~\cite{gu2018ava}, which capture a diverse set of sequences, including sports, casual interactions and movies. Our method operates on detections and masks coming from an off-the-shelf Mask-RCNN network~\cite{he2017mask}, and returns the identity label for each one of them. Therefore, the metrics we use to report results also focus on identity tracking at the level of the bounding box. More specifically, we report results using Identity switches (IDs), Multi-Object Tracking Accuracy (MOTA)~\cite{kasturi2008framework}, ID F1 score (IDF1)~\cite{ristani2016performance} and HOTA~\cite{luiten2021hota}. In all cases, we adopt the protocols of  Rajasegaran~\etal~\cite{rajasegaran2021tracking} for evaluation. 

First, we ablate the main components of our approach. Specifically, we investigate the effect of each one of the tracking cues we employ, \ie, appearance, 3D location and 3D pose, and how they affect the overall tracking pipeline. For this comparison, we report results on the Posetrack dataset~\cite{andriluka2018posetrack}. The full results are presented in Table~\ref{tbl:exp_components}. As we can see, removing each one of the main cues leads to degradation in the performance of the system, where 3D location seems to have the largest effect on the performance, followed by appearance and 3D pose. Moreover, this ablation also highlights the importance of having the nearness term in the cost function, a feature that is not available to purely 2D tracking methods. 

\begin{table}[t]
\centering
\begin{tabular}{l c c c }
\toprule[0.4mm]
\multirow{2}{*}{Method} & \multicolumn{3}{c}{PoseTrack}\\
& IDs$\downarrow$ & MOTA$\uparrow$ & IDF1$\uparrow$ \\ \midrule
w/o 3D appearance             & 632        & 58.4       & 74.9 \\
w/o 3D pose                   & 558        & 58.9       & 76.2 \\
w/o location                  & 948        & 57.3       & 71.6 \\
w/o nearness                  & 622        & 58.5       & 74.8 \\ \midrule
Full system                   & {\bf 541}  & {\bf 58.9} & {\bf 76.4} \\
\bottomrule[0.4mm]
\end{tabular}
\vspace{-0.4em}
\caption{{\bf Ablation of the main components of PHALP on PoseTrack~\cite{andriluka2018posetrack}.}
Removing each tracking cue (3D appearance, 3D pose or 3D location) leads to degradation in the performance.}
\label{tbl:exp_components}
\vspace{-0.8em}
\end{table}

Next, we evaluate our approach in comparison with the state-of-the-art methods. The results are presented in Table~\ref{tbl:exp_sota}. We report results on PoseTrack~\cite{andriluka2018posetrack}, MuPoTS~\cite{mehta2018single} and AVA~\cite{gu2018ava}. Our method outperforms the previous baselines, as well as the state-of-the-art approach of Rajasegaran~\cite{rajasegaran2021tracking}. The gains are significant across all metrics. Our method also outperforms the other approaches in the HOTA metric.

\begin{figure*}
    \centering
    \includegraphics[width=0.955\textwidth]{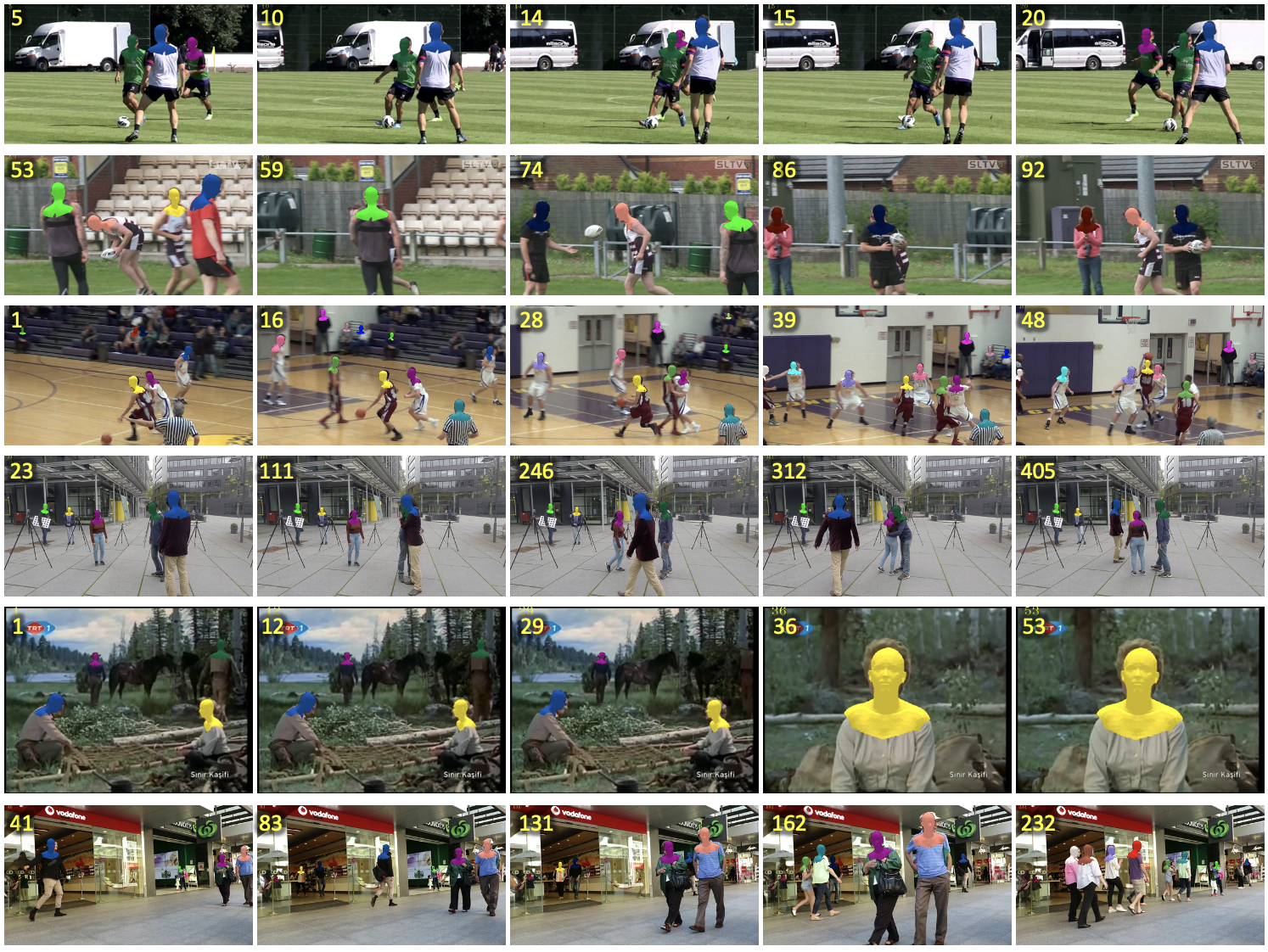}
    \caption{\textbf{Qualitative Results:} We show the tracking performance of PHALP in various datasets (frame number is shown at the top left corner).  The first three rows are from the PoseTrack dataset~\cite{andriluka2018posetrack}. These results show that even during successive occlusions our method is able to track the identity of the correct person. Note that, in the first row, although the \textcolor{OliveGreen}{green head-mask} person and the \textcolor{purple}{purple head-mask} person have similar appearance, our method can track each one of them successfully. In the second row, the \textcolor{Lavender}{player} is going through multiple occlusions, yet recovered correctly. The third row shows the robustness of our linearization approximation for 3D location prediction, even when the motions of the players are very complex. In the MuPoTS dataset~\cite{mehta2018single} (4th row), our method can handle very close interactions between people. This is due to the fact that, our modification of HMAR recovers meshes conditioned on the detected mask. We also show results (5th row) on the AVA dataset~\cite{gu2018ava}. After the 3rd frame, there is a shot change in the video, and the \textcolor{Goldenrod}{woman} is tracked successfully across the shots. Finally, we show qualitative results on a MOT17 sequence. The \textcolor{blue}{blue} person is tracked for the whole sequence while he is going through multiple occlusions for a long time. More results at the \href{https://brjathu.github.io/PHALP/}{PHALP website}.    }
    \label{fig:results_qualitative}
\end{figure*}

Finally, we also show qualitative results of our method on multiple datasets in Fig~\ref{fig:results_qualitative}. These results show that our method performs reliably even in very hard occlusion cases, while it is able to recover the correct identity over multiple successive occlusions. Fig~\ref{fig:results_qualitative} also shows the robustness of our method in complex motion sequences, shot changes and long trajectories.

\begin{table*}[!h]
\begin{center}
\begin{tabular}{l c c c c | c c c c | c c}
\toprule[0.4mm]
\multirow{2}{*}{Method} & \multicolumn{4}{c}{Posetrack} & \multicolumn{4}{c}{MuPoTS} &  \multicolumn{2}{c}{AVA}  \\
  & IDs$\downarrow$ & MOTA$\uparrow$ & IDF1$\uparrow$ & HOTA$\uparrow$ & IDs$\downarrow$ & MOTA$\uparrow$ & IDF1$\uparrow$ & HOTA$\uparrow$ & IDs$\downarrow$ & IDF1$\uparrow$\\ 
\midrule
Trackformer~\cite{meinhardt2021trackformer}       & 1263       & 33.7       & 64.0       & 46.7     & 43       & 24.9      & 62.7        & 53.2      & 716       & 40.9       \\
Tracktor~\cite{bergmann2019tracking}              & 702        & 42.4       & 65.2       & 38.5     & 53       & 51.5      & 70.9        & 50.3      & 289       & 46.8       \\
AlphaPose~\cite{fang2017rmpe}                     & 2220       & 36.9       & 66.9       & 37.6     & 117      & 37.8      & 67.6        & 41.8      & 939       & 41.9       \\
FlowPose~\cite{xiu2018pose}                       & 1047       & 15.4       & 64.2       & 38.0     & 49       & 21.4      & 67.1        & 43.0      & 452       & 52.9       \\
T3DP~\cite{rajasegaran2021tracking}               & 655        & 55.8       & 73.4       & 50.6     & 38       & 62.1      & 79.1        & 59.2      & 240       & 61.3       \\
PHALP                                             &{\bf541}    &{\bf58.9}   &{\bf76.4}   &{\bf52.9} &{\bf22}   &{\bf66.2}  &{\bf 81.4}   & \bf{59.4} &{\bf227}   &{\bf62.7}   \\
\bottomrule[0.4mm]
\end{tabular}
\end{center}
\vspace{-0.4cm}
\caption{{\bf Comparison with state-of-the-art tracking methods.} We compare our method PHALP with various tracking methods in three different datasets. Our approach outperforms the other baselines across all datasets and metrics. }
\label{tbl:exp_sota}
\end{table*}

%% file: 06_conclusion.tex
\vspace{-0.1cm}
\section{Discussion}
\vspace{-0.1cm}
We presented PHALP, an approach for monocular people tracking, by predicting appearance, location and pose in 3D. Our method relies on a powerful backbone for 3D human mesh recovery, modeling on the tracklet level for collecting information across the tracklet's detections, and eventually predicting the future states of the tracklet. One of the main benefits of PHALP is that the association aspect requires tuning of only five parameters, which makes it very friendly for training on multi-object tracking datasets, where annotating the identity of every person in a video can be expensive.  We should note that our approach can be naturally extended to make use of more attributes, \eg, a face embedding, which could be useful for cases with close-ups, like movies. The main assumptions for PHALP are that we have access to a good object detector for the initial bounding box/mask detection, and a strong HMAR network for single-frame lifting of people to 3D.  If the performance of these components is not satisfactory, it can also affect PHALP. Regarding societal impact, tracking systems have often been used for human surveillance. We do not condone such use. Instead, we believe that a tracking system will be valuable for studying social-human interactions.

{\bf Acknowledgements:} This work was supported by ONR MURI (N00014-14-1-0671), the DARPA Machine Common Sense program, as well as BAIR and BDD sponsors.